\documentclass[conference]{IEEEtran}
\usepackage{times}
\usepackage{todonotes}
\usepackage[numbers]{natbib}
\usepackage{multicol}
\usepackage{graphics,graphicx,caption,float,subcaption,booktabs,xcolor,multirow,array,color,ifthen,tabu,colortbl,dblfloatfix,url,xparse,mathtools,patchcmd,algorithm,algorithmic,amssymb,xspace,nicefrac,microtype,amsmath,amsfonts,bm,ragged2e,tikz,stackengine,etoolbox,xpatch,enumerate,xstring,setspace,tabularx,makecell,changepage,cuted,titlesec,wrapfig,tcolorbox, sidecap,comment, bbding, placeins}
\usepackage[pagebackref=true,breaklinks=true,colorlinks=true,bookmarks=false,citecolor=blue]{hyperref}
\hypersetup{
colorlinks=true,
linkcolor=blue,
filecolor=magenta,      
citecolor=blue
}

\usepackage[utf8]{inputenc}
\usepackage{textcomp}
\usepackage{array}

\IEEEoverridecommandlockouts
\begin{document}

\title{\LARGE \bf
\textit{MetaWorld-\textcolor{pink}{X}}: Hierarchical World Modeling via VLM-Orchestrated Experts for Humanoid Loco-Manipulation
}
\ifdefined\isanonymous
    \author{%
        Anonymous Authors
    \thanks{Affiliations withheld for double-blind review.
    }\\
    \thanks{
    }\\
    \thanks{
    }
    }
\else
    \author{Yutong Shen$^{*, 2}$, Hangxu Liu$^{*, 3}$, Penghui Liu$^{2}$, Jiashuo Luo$^{2}$, Yongkang Zhang$^{2}$, Rex Morvley$^{2}$, Chen Jiang$^{4}$, \\ Jianwei Zhang$^{1}$ and Lei Zhang$^{1\dag}$%
    \thanks{\dag Corresponding author. {\tt\small(zhanglei.cn.de@gmail.com)}}%
    \thanks{$^{*}$Equal contributions. {\tt\small(syt2004@emails.bjut.edu.cn)}}
        \thanks{$^{1}$University of Hamburg, Hamburg, Germany}%
    \thanks{$^{2}$School of Information Science and Technology, Beijing University of Technology, Beijing, China }%
    \thanks{$^{3}$School of Information Science and Engineering, Fudan University, Shanghai, China}%
    \thanks{$^{4}$University of Alberta, Canada}%

    }
\fi
\makeatletter
\let\@oldmaketitle\@maketitle
\renewcommand{\@maketitle}{%
  \@oldmaketitle%
  \par\vspace{1em}%
  \setcounter{figure}{0}
  \begin{center}
    \includegraphics[width=0.9\linewidth]{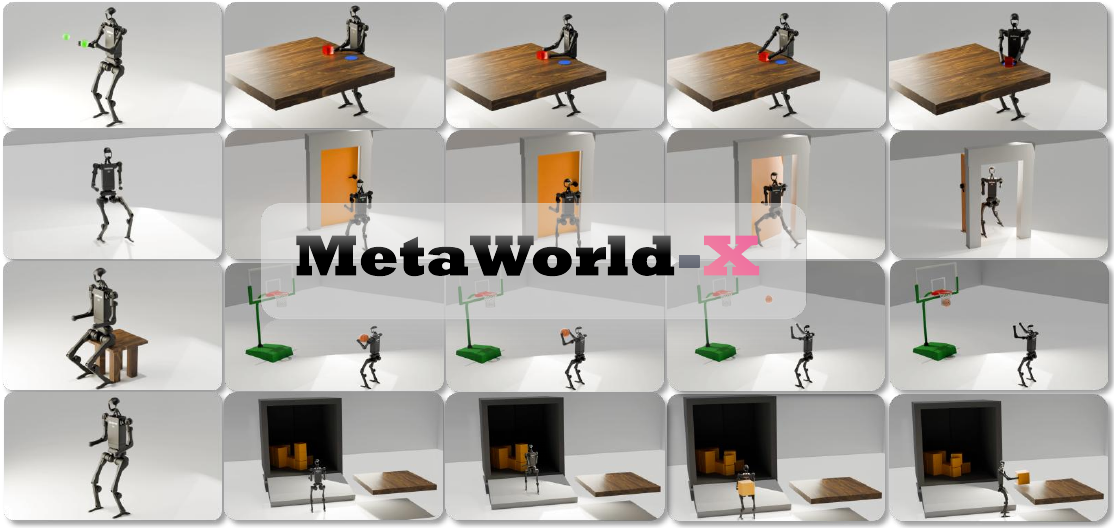}\\[0.5em]
    \captionof{figure}{
    \textbf{MetaWorld-X} is a hierarchical world model framework for whole-body humanoid control, featuring a \textbf{Specialized Expert Policy (SEP)} module for skill decomposition and an \textbf{Intelligent Routing Mechanism (IRM)} guided by a Vision-Language Model (VLM). By orchestrating a Mixture-of-Experts (MoE) architecture, it decomposes complex loco-manipulation tasks into human-motion-informed primitives and dynamically composes them via semantic routing.
    }
    \label{fig:teaser}
  \end{center}
  \vspace{-1em}%
}
\makeatother
\maketitle
\setcounter{figure}{1}
\pagestyle{empty}

\begin{abstract}

Learning natural, stable, and compositionally generalizable whole-body control policies for humanoid robots performing simultaneous locomotion and manipulation (loco-manipulation) remains a fundamental challenge in robotics. Existing reinforcement learning approaches typically rely on a single monolithic policy to acquire multiple skills, which often leads to cross-skill gradient interference and motion pattern conflicts in high-degree-of-freedom systems. As a result, generated behaviors frequently exhibit unnatural movements, limited stability, and poor generalization to complex task compositions.
To address these limitations, we propose MetaWorld-X, a hierarchical world model framework for humanoid control. Guided by a divide-and-conquer principle, our method decomposes complex control problems into a set of specialized expert policies (Specialized Expert Policies, SEP). Each expert is trained under human motion priors through imitation-constrained reinforcement learning, introducing biomechanically consistent inductive biases that ensure natural and physically plausible motion generation.
Building upon this foundation, we further develop an Intelligent Routing Mechanism (IRM) supervised by a Vision-Language Model (VLM), enabling semantic-driven expert composition. The VLM-guided router dynamically integrates expert policies according to high-level task semantics, facilitating compositional generalization and adaptive execution in multi-stage loco-manipulation tasks.
Extensive experiments on the Humanoid-bench benchmark demonstrate that MetaWorld-X significantly outperforms strong baselines in motion quality, training efficiency, and task success rates. Compared to TD-MPC2, our approach achieves substantial improvements in both return and convergence speed across multiple locomotion and manipulation tasks. These results validate the effectiveness of semantic-driven expert orchestration in mitigating multi-skill conflicts, enhancing compositional generalization, and improving motion naturalness. 
More details will be found at: \url{https://syt2004.github.io/metaworldX/}.

\end{abstract}

\IEEEpeerreviewmaketitle

\section{Introduction}

In high-degree-of-freedom (DoF) humanoid control~\cite{c4}, pure reinforcement learning (RL) methods~\cite{c1,c2,c3} offer end-to-end optimization capabilities but exhibit significant limitations in complex skill composition scenarios. In particular, when a monolithic policy architecture is employed, multiple skills share a common parameter space, which often leads to negative interference and gradient conflicts~\cite{c13}. As a result, motion patterns may become physically inconsistent or unstable during skill switching and composition. This structural deficiency prevents the policy from simultaneously maintaining balance control and precise manipulation. In complex loco-manipulation tasks, such issues typically manifest as motion jitter, unnatural gait patterns, or posture collapse.

In recent years, world model-based RL~\cite{c28} has emerged as a prominent paradigm for continuous control. These methods learn a latent representation of environment dynamics and perform trajectory prediction and planning within the latent space, thereby improving sample efficiency and enabling foresighted decision-making. 
Representative methods such as TD-MPC2~\cite{c14} learn an implicit world model and conduct model predictive control in latent space. DreamerV3~\cite{c27} learns a latent dynamics model directly from pixel observations and optimizes behavior through imagined rollouts.

However, in high-DoF humanoid control, world model approaches remain fundamentally limited. Accumulated model bias in long-horizon rollouts induces value overestimation and policy mismatch, while monolithic architectures preserve cross-skill gradient interference. Moreover, optimizing primarily for task return rather than biomechanical consistency often yields unnatural or unstable motions.

In contrast, incorporating human motion priors~\cite{c29} introduces biomechanically grounded inductive biases into policy learning, regularizing motion generation toward smoother and more natural behaviors~\cite{c6}. However, such priors entail structural retargeting discrepancies, human–robot dynamical mismatches, and objective conflicts between imitation and task optimization.

To alleviate gradient interference in monolithic policies, Mixture-of-Experts (MoE)~\cite{c5} and modular learning have gained traction in robotic control. However, prior MoE approaches primarily emphasize skill diversification rather than semantic composition for long-horizon structured tasks~\cite{c7}. Lacking explicit semantic grounding, expert combination is often heuristic or shallowly learned, making structural decomposition alone insufficient for stable, natural, and generalizable multi-stage loco-manipulation.

To address these limitations, we propose MetaWorld-X, a hierarchical control framework that integrates world model representations, human motion priors, and semantic-driven expert composition. Specifically, our method leverages world models to enhance data efficiency and planning capability and trains specialized experts under human motion priors to ensure biomechanical naturalness. As well, it employs a VLM-guided semantic routing mechanism to enable compositional generalization. By combining structural modularization with semantic orchestration, our framework resolves skill conflicts at the architectural level while coordinating naturalness and task objectives at the optimization level.

Our primary contributions are:
\begin{enumerate}
   \item \textbf{Human-Motion-Informed Specialized Expert Learning.} We propose a human-motion-informed SEP framework that formulates humanoid control as imitation-constrained RL. By introducing an energy-based alignment reward with dynamic reweighting and a principled motion retargeting pipeline, our approach enables skill-specialized expert policies to learn biomechanically consistent and natural motion primitives while mitigating cross-skill gradient interference.

    \item \textbf{VLM-Guided Semantic Routing for Compositional Control.} We develop a VLM-supervised IRM that embeds semantic reasoning into the control loop. The router decomposes complex tasks into expert compositions, and a few-shot semantic transfer strategy allows the router to learn expert weight distributions from limited demonstrations, enabling smooth transition from supervised guidance to autonomous decision-making and supporting zero-shot compositional generalization.

   \item \textbf{Substantial Performance and Generalization Gains.} Extensive experiments on Humanoid-bench demonstrate that MetaWorld-X consistently outperforms strong state-of-the-art baselines in terms of return, sample efficiency, and motion quality. Our method achieves improvements over TD-MPC2 across locomotion and manipulation tasks, while enabling few-shot compositional generalization. Ablation studies further confirm the complementary roles of SEP and IRM in enhancing efficiency, stability, and semantic grounding.
\end{enumerate}

\section{Related Works}

\begin{figure*}[t]
    \centering
    \includegraphics[width=0.95\textwidth]{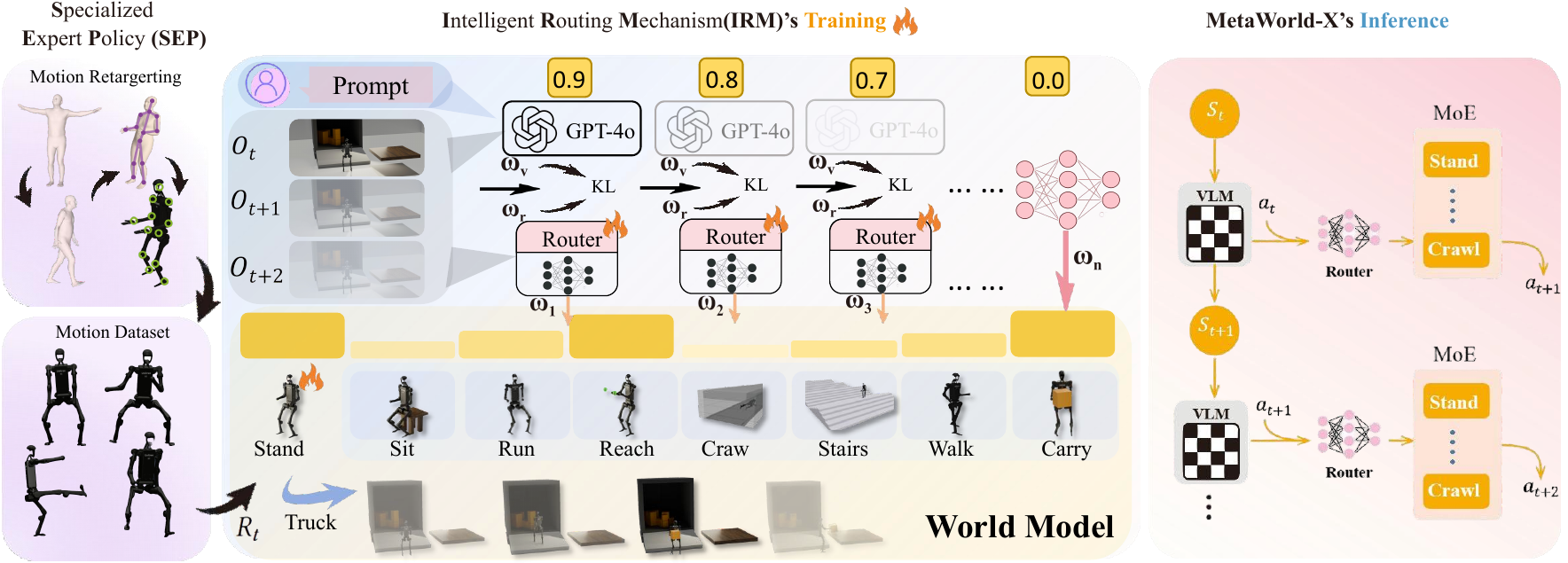}
    \caption{\textbf{MetaWorld-X} achieves natural humanoid control through the dynamic orchestration of expert policies guided by a Vision-Language Model (VLM). The framework consists of two core modules: the \textbf{Skill Expert Pool (SEP)} trains specialized policy networks for fundamental motor skills using human motion data, while the \textbf{Intelligent Router Module (IRM)} employs a VLM as a supervisory teacher. The VLM guides the router’s training via few-shot inference, enabling a seamless transition from supervised learning to autonomous operation.
}
    \label{fig.2}
\end{figure*}

\subsection{Humanoid Locomotion and Manipulation}

Achieving fluid and stable whole-body control for humanoids is a long-standing goal. This "loco-manipulation" capability presents immense technical hurdles due to the high dimensionality and complex dynamics of humanoid platforms. 
Traditional Model Predictive Control~\cite{c8,c9} and Whole-Body Control~\cite{c10,c11} rely on accurate dynamic models~\cite{c12} but struggle with scalability and generalization in unstructured environments. 
Their performance is fundamentally tied to model accuracy, which is difficult to maintain in the real world. Learning-based approaches, particularly those using a single, monolithic policy trained with Reinforcement Learning (RL), have emerged as a powerful alternative. These methods learn complex behaviors directly from interaction. 

\subsection{Reinforcement Learning for Humanoid}
Within learning-based methods, model-based RL algorithms have shown great promise for continuous control tasks due to their sample efficiency and planning capabilities~\cite{c14,c16}. TD-MPC2 learns an implicit world model and performs trajectory optimization in its latent space. While powerful, its application to very high-DoF humanoids has revealed challenges, including value overestimation and policy mismatch, which can degrade performance in complex locomotion tasks~\cite{c15}.

\subsection{Modularity, Semantics, and Imitation}
\textbf{Mixture-of-Experts (MoE) in Robotics}: The MoE architecture combats negative interference by routing inputs to specialized experts. Prior works have demonstrated MoE for learning diverse skills. We uniquely use MoE to manage a library of locomotion and manipulation skills for true compositionality~\cite{c17,c18}. 

\textbf{Vision-Language Models (VLMs) for Task Understanding}: VLMs ground language in visual data for robot task planning. Web-scale knowledge has been shown to transfer to robotic control~\cite{c19,c20}. However, a semantic gap often exists between high-level plans and execution. 
We differ by using the VLM as a supervisory signal for expert routing.

\textbf{Imitation Learning (IL) for Natural Motion}: Achieving human-like motion is critical for social acceptance~\cite{c21,c22,c23}. Pure RL often yields jerky gaits. Imitation learning from motion capture data (e.g., AMASS) infuses naturalness into policies. A key limitation is that IL for a single skill does not inherently solve composition for complex tasks. MetaWorld-X addresses this by using IL to train individual experts on natural motion data, ensuring a high-quality foundation for composition.

\section{Methodology}

\subsection{Overview of MetaWorld-X}
\begin{figure*}[t]
    \centering
    \includegraphics[width=0.9\textwidth]{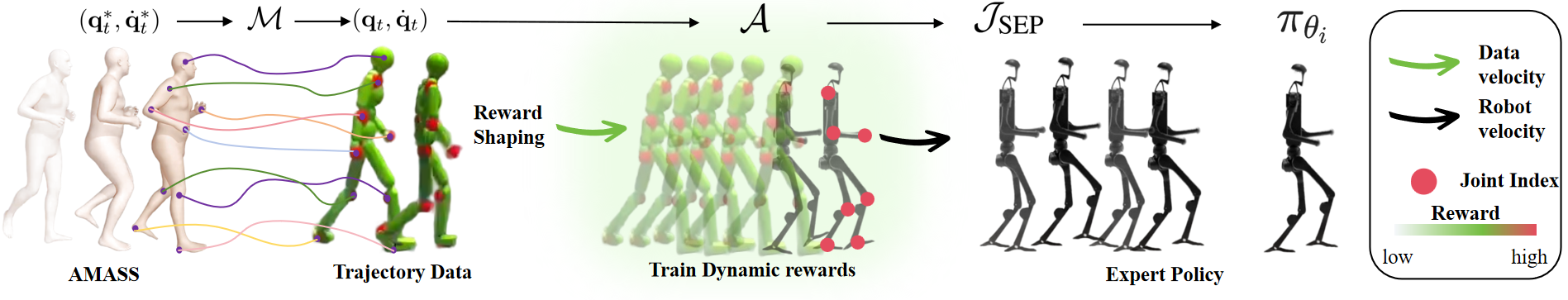}
    \caption{\textbf{Architecture of the \textbf{SEP} module.} We project human motion priors to the robot's configuration space via operator $\mathcal{M}$, then utilize alignment operator $\mathcal{A}$ to compute tracking signals for expert policies $\pi_{\theta_i}$. By optimizing $\mathcal{J}_{\text{SEP}}$ with a dynamic weighting mechanism that prioritizes poorly-tracked joints, the module yields the converged expert policies.}
    \label{fig.3}
\end{figure*}

\textbf{MetaWorld-X} is a hierarchical world model framework for humanoid loco-manipulation, as shown in Fig.~\ref{fig.2}. It employs a ‘divide-and-conquer’ strategy, decomposing complex tasks into motion primitives managed by Specialized Expert Policies (SEP) to mitigate skill interference, as shown in Fig.~\ref{fig.3}. We formulate the objective of the SEP module as:
\begin{equation}
    \mathcal{J}_{\text{SEP}} = \sum_{i=1}^{K} \mathbb{E}_{s \sim \mathcal{D}_{\text{human}}} \left[ \mathcal{A}(\pi_i(s), \pi_{\text{human}}^{(i)}) \right]
\end{equation}
where $\mathcal{A}$ denotes the alignment operator, aiming to maximize the fidelity of $K$ expert policies $\{\pi_i\}_{i=1}^K$ relative to human motion priors $\pi_{\text{human}}^{(i)}$, ensuring biomechanically plausible motor primitives.

Orchestrated by a VLM-supervised Intelligent Routing Mechanism (IRM) leveraging human motion priors, the framework dynamically dispatches expert policies to achieve complex task goals while enhancing generalization. The IRM objective is defined by the semantic composition:
\begin{equation}
    \pi_{\text{mix}}(s) = \mathcal{F}_{\phi} \left( \{ \pi_i(s) \}_{i=1}^K ; \mathcal{V}(\mathcal{T}) \right)
\end{equation}
Here, $\mathcal{F}_{\phi}$ represents the composition operator parameterized by $\phi$, which integrates expert policies based on the semantic guidance $\mathcal{V}(\mathcal{T})$ provided by the VLM and the $\mathcal{T}$ means tasks. This enables dynamic, context-aware skill synthesis, allowing the framework to transition beyond fixed primitives toward autonomous task execution.

\subsection{Specialized Expert Policy Learning}
\label{111}
The SEP module optimizes motion primitives by framing humanoid control as an imitation-constrained RL problem, aiming to maximize the cumulative energy-based alignment along the expert trajectory to stabilize imitation under imperfect retargeting. The alignment operator $\mathcal{A}$ is formulated:

\begin{equation}\label{eq:sep_objective}
    \mathcal{J}_{SEP} = \mathbb{E}_{\pi} \left[ \sum_{t=0}^{T} \gamma^t (R_t + \beta \mathcal{H}(\pi(\cdot|s_t))) \right]
\end{equation}

where $\gamma \in (0, 1]$ is the discount factor, $R_t$ denotes the energy-based imitation reward (the temporal implementation of $\mathcal{A}$), $\mathcal{H}(\pi) = -\mathbb{E}_{\pi}[\log \pi(\cdot|s_t)]$ represents the policy entropy regularization, and $\beta$ is the temperature coefficient.

\subsubsection{Energy-Based Imitation Reward}
SEP translates expert priors into a dense, dynamic reward by treating imitation as energy minimization. The instantaneous reward $R_t$ (as the instantiation of $\mathcal{A}$) is an exponential alignment between the robot's state and the reference expert trajectory:

\begin{equation}\label{eq:reward}
R_t = w \left[ e^{-\alpha \| \mathbf{q}_t - \mathbf{q}^*_t \|^2} + \lambda e^{-\beta \| \dot{\mathbf{q}}_t - \dot{\mathbf{q}}^*_t \|^2} \right]
\end{equation}

where $(\mathbf{q}_t, \dot{\mathbf{q}}_t)$ and $(\mathbf{q}^*_t, \dot{\mathbf{q}}^*_t)$ are the robot and reference expert states at phase-aligned time $t$, and $w, \alpha, \beta, \lambda$ are scalar hyperparameters for reward shaping. 
This corresponds to maximizing the log-likelihood under a Gaussian assumption.
The exponential form yields bounded, non-vanishing gradients, ensuring stable learning near and far from the target.

\subsubsection{Dynamic Reward Reweighting}
This dense reward, propagating the local alignment signal of $\mathcal{A}$, is integrated into the world model's reward head $\hat{\mathcal{R}}_\theta(z_t, a_t)$.  
The alignment-based value head, denoted by $Q_{\mathcal{A}}$ (parameterized by $\bar{\theta}$), learns via distributional TD:

\begin{equation}\label{eq:td_update}
y_t = R_t + \gamma Q_{\mathcal{A}}(z_{t+1}, \pi_\theta(z_{t+1}))
\end{equation}

propagating the alignment signal forward. Policy updates follow a maximum-entropy objective:

\begin{equation}\label{eq:policy_update}
J(\pi) = \mathbb{E}\left[ Q_{\mathcal{A}}(z, a) - \tau \mathcal{H}(\pi(\cdot|z)) \right]
\end{equation}

where $\tau$ is the entropy weight, $Q_{\mathcal{A}}$ is the alignment-based action-value function, and $\mathcal{H}$ is the entropy operator, balancing imitation-driven return with exploration to avoid covariate shift. For planning, MPPI/CEM optimizes in latent space over horizon $H$:

\begin{equation}\label{eq:planning_objective}
J = \sum_{k=0}^{H-1} \gamma^k \left( \hat{\mathcal{R}}_\theta(z_{t+k}, a_{t+k}) + \mu Q_{\mathcal{A}}(z_{t+k}, a_{t+k}) \right)
\end{equation}

where $\mu$ is the scaling factor for the value head. This frames imitation energy as a stage cost, enabling optimal control with style consistency. Latent planning with dynamics regularization allows internal simulation of contact and acceleration effects on future tracking, facilitating rapid recovery.

\subsection{Human Motion Prior and Retargeting}

We adapt MoCap data to our humanoid using the H2O framework~\cite{c24}. The SMPL model is aligned to the robot morphology via shape optimization, followed by motion retargeting formulated as a state-alignment operator:
\begin{equation}
\mathcal{M}(\mathbf{q}_t, \dot{\mathbf{q}}_t, \mathbf{q}^*_t, \dot{\mathbf{q}}^*_t)
= \| \mathbf{q}_t - \mathbf{q}^*_t \|^2_W 
+ \gamma \| \dot{\mathbf{q}}_t - \dot{\mathbf{q}}^*_t \|^2_W .
\end{equation}
Inverse kinematics minimizes $\mathcal{M}$ under physical and joint constraints. 
Dynamic feasibility is verified in simulation, where unstable trajectories are filtered via a privileged imitation policy~\cite{c25}, yielding a refined motion dataset for expert training.

\subsection{VLM-Guided Semantic Routing for Expert Composition}

\vspace{-10pt}
\begin{figure}[hbpt]
    \centering
    \includegraphics[width=0.45\textwidth]{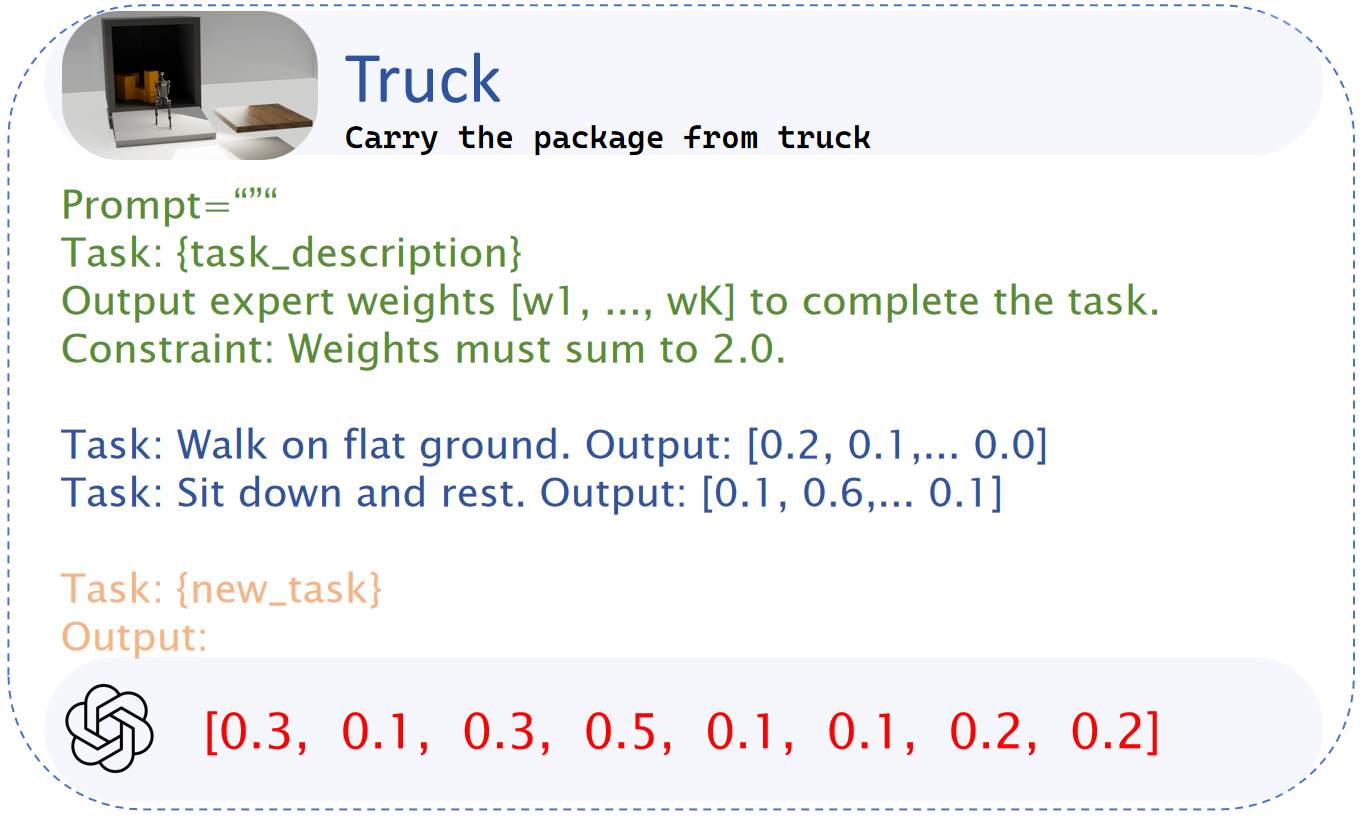}
    \caption{Example of VLM prompt for Semantic Router.}
        \label{fig.prompt}

\end{figure}
\vspace{0pt}

\begin{table*}[t]
\centering
\small
\caption{Expert policy evaluation: Our \textbf{IRM} module vs. baselines. We report maximum returns ($R. \pm stderr.$) and convergence steps $S$ in Millions (M). $\uparrow$ indicates higher is better, $\downarrow$ indicates lower is better.}
\label{tab:results}
\vspace{2mm}
\resizebox{0.9\textwidth}{!}{%
\begin{tabular}{@{} l cc cc cc cc cc @{}} 
\toprule
\multirow{2}{*}{\textbf{Algorithm}} & \multicolumn{2}{c}{\textbf{Stand}} & \multicolumn{2}{c}{\textbf{Walk}} & \multicolumn{2}{c}{\textbf{Run}} & \multicolumn{2}{c}{\textbf{Sit}} & \multicolumn{2}{c}{\textbf{Carry}} \\
\cmidrule(lr){2-3} \cmidrule(lr){4-5} \cmidrule(lr){6-7} \cmidrule(lr){8-9} \cmidrule(lr){10-11}
& R.$\pm$stderr. & S.(M) & R.$\pm$stderr. & S.(M) & R.$\pm$stderr. & S.(M) & R.$\pm$stderr. & S.(M) & R.$\pm$stderr. & S.(M) \\
\midrule
PPO~\cite{c25}        & 28.8$\pm$0.4   & 5.3 & 28.8$\pm$0.4    & 5.3 & 28.8$\pm$0.4   & 5.3 & 28.8$\pm$0.4    & 5.3 & 25.9$\pm$0.4   & 5.4 \\
SAC ~\cite{c26}      & 93.6$\pm$17.7  & 4.9 & 32.7$\pm$11.4   & 5.4 & 18.3$\pm$1.5   & 5.4 & 275.9$\pm$15.7  & 6.0 & 43.4$\pm$9.7   & 5.3 \\
TD-MPC2~\cite{c14}   & 749.8$\pm$63.1 & 1.8 & 644.2$\pm$162.3 & 1.8 & 66.1$\pm$4.7   & 2.0 & 733.9$\pm$120.6 & 1.1 & 438.0$\pm$72.9 & 1.9 \\
DreamerV3~\cite{c27} & 699.3$\pm$62.7 & 5.5 & 428.2$\pm$14.5  & 6.0 & 298.5$\pm$84.5 & 6.0 & 709.6$\pm$98.4  & 5.2 & 427.8$\pm$51.2 & 6.0 \\
\midrule
\textbf{Ours (IRM)} & \textbf{815.9$\pm$0.3$\uparrow$} & \textbf{0.6$\downarrow$} & \textbf{1118.7$\pm$7.1$\uparrow$} & \textbf{0.5$\downarrow$} & \textbf{2056.9$\pm$13.6$\uparrow$} & \textbf{1.0$\downarrow$} & \textbf{862.2$\pm$2.1$\uparrow$} & \textbf{0.6$\downarrow$} & \textbf{963.5$\pm$5.1$\uparrow$} & \textbf{0.5$\downarrow$} \\
\bottomrule
\end{tabular}
} %
\end{table*}

\vspace*{-10pt}
\begin{figure*}[t]
\vspace*{-1pt}
    \centering
    \includegraphics[width=0.9\textwidth]{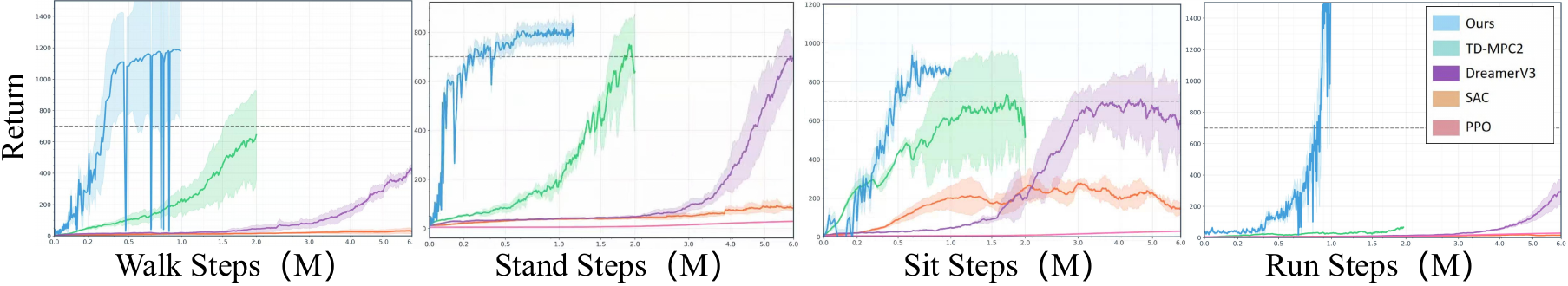}
    \caption{We evaluate the performance of four foundational motor skills: \textit{Walking, Running, Standing, and Sitting}, after 50w training steps. The dashed lines in the learning curves qualitatively represent the reward thresholds associated with the successful execution of each skill.}
    \label{fig:exp1}
\end{figure*}

\vspace{1em}
\begin{figure}[t]
    \centering
    \includegraphics[width=\linewidth]{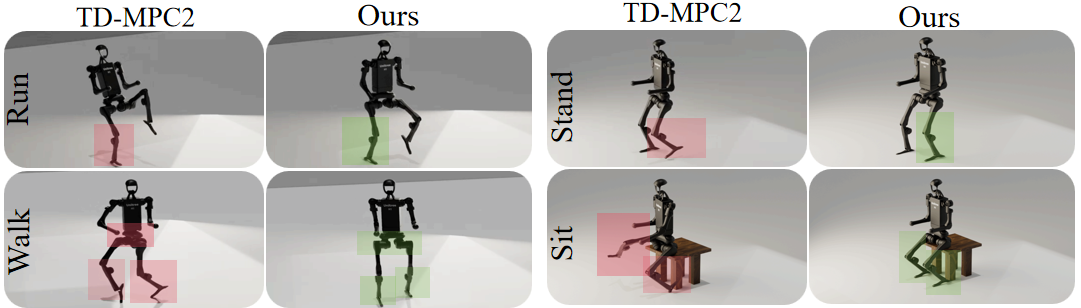}
    \caption{In comparison with TD-MPC2, Green shaded areas highlight the superior performance of \textbf{Ours}, whereas red regions indicate limitations observed in baseline algorithms.}
    \label{fig:compare-results-tdmpc2-and-ours}
\end{figure}

To operationalize the composition operator $\mathcal{F}_{\phi}$, we propose an IRM that synthesizes motor primitives through a learnable routing network $\pi_{\phi}$. Given $K$ expert policies $\{\pi_i\}_{i=1}^K$ and the VLM-provided semantic guidance $\mathcal{V}(\mathcal{T})$, the routing network $\pi_{\phi}(\mathbf{w}|s, \mathcal{V}(\mathcal{T}))$ maps the current observation $s$ to a weight distribution $\mathbf{w} \in \Delta^{K-1}$. The composition is defined as:
\begin{equation}
\pi_{\text{mix}}(s) = \mathcal{F}_{\phi}(\{\pi_i(s)\}_{i=1}^K ; \mathcal{V}(\mathcal{T})) = \sum_{i=1}^K w_i \cdot \pi_i(s)
\end{equation}
where $w_i = [\pi_{\phi}(\mathbf{w}|s, \mathcal{V}(\mathcal{T}))]_i$ denotes the weight assigned to the $i$-th expert. To achieve autonomous task execution, we propose a hierarchical semantic distillation pipeline:

\textbf{Task-Level Semantic Prior (Coarse Alignment).}
We derive task-level relevance scores $\mathbf{w}_v \in \mathbb{R}^K$ from the VLM based on the task description $\mathcal{T}$. The router aligns with this global guidance to distill foundational task semantics:
\begin{equation}
\mathcal{L}_{\text{task}}(\phi) = \mathcal{D}_{\text{KL}}(\pi_{\phi}(\mathbf{w}|s, \mathcal{V}(\mathcal{T})) \, \| \, \mathbf{w}_v) - \beta \mathcal{H}(\pi_{\phi})
\end{equation}
where $\mathbf{w}_v$ represents the zero-shot expert relevance inferred by the VLM for task $\mathcal{T}$, $\beta$ is the entropy regularization coefficient, and $\mathcal{H}(\cdot)$ is the policy entropy. This objective determines which experts are globally relevant for a given task, providing essential semantic grounding.

\textbf{Demonstration-Level Behavioral Refinement (Fine Alignment).}
To capture temporal composition, we use few-shot demonstrations $D$. The VLM extracts a demonstration-conditioned prior $P_{\text{demo}}(\mathbf{w}|D) \in \Delta^{K-1}$, where $P_{\text{demo}}$ reflects expert usage statistics observed in trajectories. The router is optimized via:
\begin{equation}
\mathcal{L}_{\text{demo}}(\phi) = \mathbb{E}_{s \sim \mathcal{D}_{\mathcal{T}}} \left[ \mathcal{D}_{\text{KL}} \left( \pi_{\phi}(\mathbf{w}|s, \mathcal{V}(\mathcal{T})) \, \| \, P_{\text{demo}}(\mathbf{w}|D) \right) \right]
\end{equation}
Unlike $\mathcal{L}_{\text{task}}$, which provides coarse guidance, $\mathcal{L}_{\text{demo}}$ refines the router to match the temporally consistent expert composition statistics found in demonstrations $D$, as shown in Fig.~\ref{fig.prompt}.

\textbf{Unified Training Objective.}
The final objective integrates coarse task semantics and fine-grained behavioral priors into a unified hierarchical supervision framework:
\begin{equation}
\mathcal{L}_{\text{IRM}}(\phi) = \lambda(t) \cdot \mathcal{L}_{\text{task}}(\phi) + \mathcal{L}_{\text{demo}}(\phi)
\end{equation}
where $\lambda(t) = \lambda_0 \eta^t$ is a time-decaying guidance weight, with $\lambda_0$ as the initial weight, $\eta \in (0,1)$ as the decay rate, and $t$ denoting the training iteration. This schedule facilitates a transition from VLM-guided bootstrapping (high $\lambda(t)$) to demonstration-based behavioral refinement (dominant $\mathcal{L}_{\text{demo}}$). Task-level alignment provides semantic grounding, while few-shot transfer refines behavioral composition. Together, they form a hierarchical semantic distillation mechanism for expert routing.

\subsection{Inference Procedure}
At inference time, the IRM module functions as a low-latency routing network that synthesizes motor primitives in real-time. Given the cached task guidance $\mathcal{V}(\mathcal{T})$ and the current observation $s_t$, the routing network produces the optimal weight distribution $\mathbf{w}_t = \pi_\phi(s_t, \mathcal{V}(\mathcal{T}))$, enabling the composition operator to generate the final action:
\begin{equation}
a_t = \mathcal{F}_{\phi}(\{\pi_i(s_t)\}_{i=1}^K ; \mathbf{w}_t) = \sum_{i=1}^K w_{t,i} \cdot \pi_i(s_t)
\end{equation}
This avoids redundant VLM queries during control, ensuring the framework maintains high-frequency execution for complex loco-manipulation tasks.

\vspace{1em}

\section{Experiment}
To validate our approach, we conduct comparison experiments on both fundamental locomotion tasks and complex locomotion and manipulation tasks from the Humanoid-bench~\cite{c13}, ablation study.

\subsection{Comparison Experiments of Policy Learning}
\textbf{Experimental Setup.}~
Our primary objective is to leverage expert policies for basic skills to accelerate learning in complex locomotion and manipulation tasks. To this end, we construct eight minimally orthogonal robotic skill experts: \textit{Stand, Walk, Run, Sit, Carry, Reach, and Crawl}.

\begin{table}[t]
\centering
\small %
\caption{Performance comparison on locomotion tasks (Success Rate / 10). \textbf{Ours} outperforms all baseline RL and model-based algorithms.}
\label{tab:success_rate_locomotion}
  \resizebox{0.8\linewidth}{!}{
\begin{tabular}{ l ccccc }
\toprule
\textbf{Method} & \textbf{Stand} & \textbf{Walk} & \textbf{Run} & \textbf{Sit} & \textbf{Carry} \\
\midrule
PPO~\cite{c25}             & 0/10          & 0/10          & 0/10          & 0/10          & 0/10          \\
SAC~\cite{c26}             & 1/10          & 1/10          & 0/10          & 2/10          & 1/10          \\
DreamerV3~\cite{c27}       & 2/10          & 2/10          & 1/10          & 3/10          & 2/10          \\
TD-MPC2~\cite{c14}         & 3/10          & 3/10          & 2/10          & 4/10          & 3/10          \\
\midrule
\textbf{Ours}   & \textbf{9/10} & \textbf{9/10} & \textbf{9/10} & \textbf{8/10} & \textbf{9/10} \\
\bottomrule
\end{tabular}
}
\end{table}

\begin{figure}[t]
    \centering
    \includegraphics[width=0.80\linewidth]{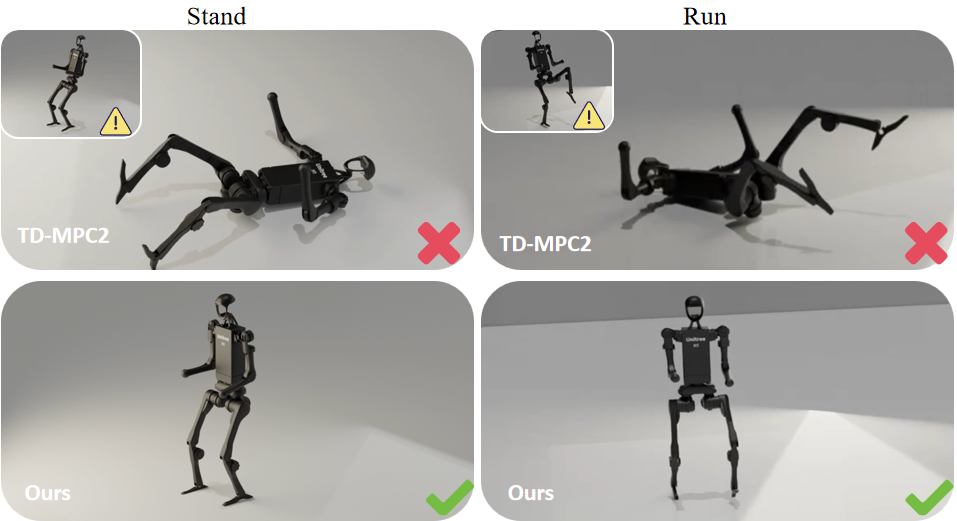}
    \caption{We evaluated the success rates of our method and Others on basic expert skills over 500k training iterations. For each task, ten independent trials were conducted. A trial was considered successful if the agent did not fall within a 30‑second episode and complete the task. The results demonstrate that after 500k steps, our method achieves a significantly higher success rate than Others. This confirms the robustness and reliability of our expert‑policy learning in these locomotion tasks.}
    \label{fig:SR}
\end{figure}

\begin{figure}[t]
    \centering
    \includegraphics[width=\linewidth]{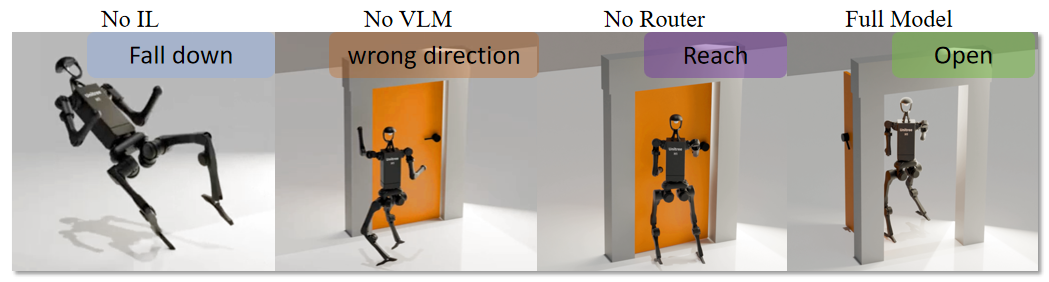}
    \caption{The effectiveness of each ablated method on the door-opening task is illustrated in this figure.}
    \label{fig:ablation}
\end{figure}

\begin{table}[t]
  \centering
  \label{tab:ablation_door}
  \caption{Among the ablated variants, the model without Imitation Learning (IL) exhibits poor performance, primarily due to its incompatibility with our subsequent training and adaptation pipeline. Meanwhile, the without Vision-Language Model (VLM) configuration is not trained, as we aim to directly employ a rule-based router to achieve zero-shot compositional generalization without semantic guidance.}
   \resizebox{0.8\linewidth}{!}{
  \begin{tabular*}{\linewidth}{l @{\extracolsep{\fill}} ccc}
    \toprule
    \textbf{Method} & \textbf{Suceess} & \textbf{steps(w)} & \textbf{return} \\
    \midrule
    TD-MPC2~\cite{c14}    & $\checkmark$ & 32.38  & 198.42 \\
    w/o Router & $\checkmark$ & 20.36  & 296.57 \\
    w/o VLM    & $\times$     & NULL   & NULL   \\
    w/o IL     & $\times$     & $\infty$  & 193.61 \\
    \textbf{Full Model} & $\checkmark$ & \textbf{12.64}  & \textbf{303.95} \\
    \bottomrule
  \end{tabular*}
  }
  \label{tab:ablation_door}
\end{table}

\begin{figure*}[t]
\vspace*{-1pt}
    \centering
    \includegraphics[width=0.9\textwidth]{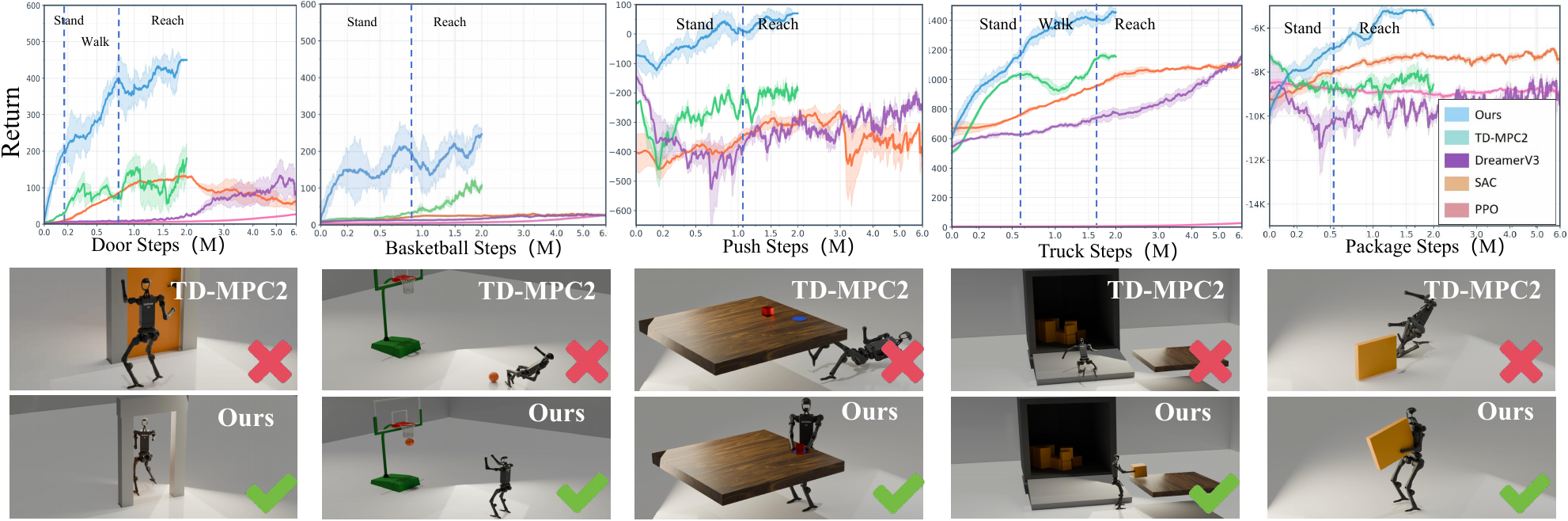}
    \caption{This figure illustrates a performance comparison between \textbf{Ours} and several baselines across various manipulation tasks. The plot includes a "Reward Switcher" curve, which indicates the specific reference expert policy being utilized for reinforcement learning at each stage of the training process.}

    \label{fig:exp3}
\end{figure*}

\begin{table*}[t]
\centering
\caption{Performance comparison on manipulation tasks. Routing schedules are indicated in Million steps (M) under each task, where \textbf{S}, \textbf{W}, and \textbf{R} denote \textit{Stand}, \textit{Walk}, and \textit{Reach} reference stages, respectively.}
\label{tab:performance_compact}
\resizebox{0.9\linewidth}{!}{
\begin{tabular}{l m{3.5cm} c c c c c}
\toprule
\textbf{Task} & \textbf{Router (M)} & \textbf{PPO}~\cite{c25} & \textbf{SAC}~\cite{c26} & \textbf{TD-MPC2}~\cite{c14} & \textbf{DreamerV3}~\cite{c27} & \textbf{Ours} \\ 
\midrule
\textbf{Door} & \scriptsize 0.17(S) $\to$ 0.74(W) $\to$ 2.0(R) & $58.0 \pm 13.5$ & $168.7 \pm 3.9$ & $285.0 \pm 12.0$ & $257.7 \pm 37.9$ & $\mathbf{470.0 \pm 2.2}$ \\
\midrule
\textbf{Basketball} & \scriptsize 0.87(S) $\to$ 2.0(R) & $58.0 \pm 1.3$ & $45.4 \pm 5.5$ & $148.4 \pm 3.3$ & $77.6 \pm 1.2$ & $\mathbf{250.0 \pm 11.9}$ \\
\midrule
\textbf{Push} & \scriptsize 1.09(S) $\to$ 2.0(R) & $36.8 \pm 11.0$ & $-201.3 \pm 13.6$ & $-113.8 \pm 6.8$ & $-59.9 \pm 7.3$ & $\mathbf{70.0 \pm 2.1}$ \\
\midrule
\textbf{Truck} & \scriptsize 0.57(S) $\to$ 1.61(W) $\to$ 2.0(R) & $58.0 \pm 13.5$ & $1142.6 \pm 15.8$ & $1213.2 \pm 1.1$ & $1383.4 \pm 2.5$ & $\mathbf{1500.0 \pm 15.6}$ \\
\midrule
\textbf{Package} & \scriptsize 0.51(S) $\to$ 2.0(R) & $-8112.3 \pm 191.1$ & $-6662.1 \pm 137.2$ & $-6788.5 \pm 552.7$ & $-5817.2 \pm 903.8$ & $\mathbf{-5200.0 \pm 47.2}$ \\
\bottomrule
\end{tabular}
}
\end{table*}

\textbf{Baselines.}~
To evaluate the effectiveness of our method, we compare against standard model-free algorithms such as PPO~\cite{c24} and SAC~\cite{c26}  as reference baselines. 
DreamerV3~\cite{c27} learns a latent world model from pixels and performs imagination-based planning, achieving strong data efficiency. 
TD-MPC2~\cite{c14} is a model-based RL framework that integrates world model learning with trajectory optimization. 

\textbf{Metrics.} We evaluate each expert policy using three complementary metrics. First, \textbf{peak return} captures the maximum average performance achieved under a fixed training budget across multiple independent random seeds, defining the method's performance ceiling. We quantify the uncertainty of the sample mean via the \textbf{Standard Error (Stderr)}, calculated as $\text{Stderr} = s/\sqrt{n}$, where $s$ denotes the sample standard deviation and $n$ represents the number of independent seeds. Finally, \textbf{convergence step} measures training efficiency as the minimum training iteration where performance enters and remains stable within a $\pm 5\%$ tolerance band of the final value for a duration of $W$ consecutive evaluations. 

\textbf{Implementation Details.} Our environment configurations follow Humanoid-bench~\cite{c13}. For the SEP module, we employ 2-layer MLPs (256 units/layer) as encoders mapping observations to a 512-dimensional latent space. Both world model and policy are 512-dimensional MLPs, while the value function utilizes an ensemble of five independent 512-dimensional Q-networks. Models support variable input dimensions (51 or 151) and action spaces (19 or 61). Experiments are conducted in MuJoCo, with performance measured via returns and standard deviation across multiple seeds.

\textbf{Results.}~
We evaluate the effectiveness and robustness of our expert policy learning, with results detailed in Tab.~\ref{tab:results} and Tab.~\ref{tab:success_rate_locomotion}. 

As shown in Fig.~\ref{fig:exp1}, our method outperforms baselines including TD-MPC2~\cite{c14} and DreamerV3~\cite{c27}. While TD-MPC2 and DreamerV3 exhibit significantly slower convergence and lower returns. Notably, our approach achieves the highest returns across all five locomotion tasks with superior sample efficiency, particularly in walking and running.

To assess kinematic naturalness, we perform a qualitative visual analysis. As illustrated in Fig.~\ref{fig:compare-results-tdmpc2-and-ours} and Fig.~\ref{fig:SR}, baseline methods often exhibit coordination deficiencies such as excessive arm swing, unstable hip and head orientation, and unnatural leg lifting. In contrast, our method effectively ameliorates these artifacts, generating gait patterns that closely resemble biological locomotion. These results demonstrate that our method successfully balances task performance with motion naturalness, establishing a robust foundation for downstream policy transfer.

\subsection{Complex Loco-Manipulation Task Evaluation Experiments.}
\textbf{Experimental Setup.}
In this section, we evaluate the performance of the intelligent router in complex loco-manipulation tasks, assess task completion, and examine the generalization capability of foundational skills to these complex scenarios, aiming to achieve more efficient and stable learning. To ensure comprehensive coverage of diverse settings, we selected representative tasks including \textbf{Basketball} for rapid dynamic motion, \textbf{Push} for static table-top manipulation, \textbf{Package} and \textbf{Truck} for whole-body coordinated manipulation, and \textbf{Door} for long-horizon sequential tasks. Across all tasks, we employ a VLM to assist in training the intelligent router and directly utilize a pre-trained router to evaluate task completion.

\textbf{Result.}
 As shown in Tab.~\ref{tab:performance_compact} and figure~\ref{fig:exp3}, Our method significantly outperforms all baselines. The improvement stems from our SEP, which mitigates ``skill gradient interference'' by decoupling motor primitives such as \textit{Stand}, \textit{Walk}, and \textit{Reach}. This separation prevents physical conflicts between balance control and precision manipulation. Furthermore, the VLM-driven semantic routing mechanism leverages multimodal priors to execute precise policy switching at critical nodes (e.g., $0.17$M or $0.57$M steps). This allows the humanoid to transform complex instructions into coherent action chains with high sample efficiency. Additionally, by incorporating AMASS motion priors, our method eliminates the motion artifacts prevalent in baselines during Push and Package tasks, ensuring both high cumulative returns and biomechanical stability.

\subsection{Ablation Study}
\textbf{Experimental Setup.}
To evaluate the contribution of each component, we conduct an ablation study on the door-opening task, with success defined by task completion. We compare five configurations trained for 500,000 steps:  the \textbf{Full model} integrates all proposed components and the complete training pipeline; the \textbf{Without intelligent router} variant replaces the learnable router with direct VLM-based expert weight referencing; the \textbf{Without VLM semantic guidance} variant replaces VLM guidance with a rule-based routing mechanism; the \textbf{Without imitation learning (IL)} variant bypasses the IL stage by mapping network dimensions directly to TD-MPC2 weights; and the \textbf{Baseline (scratch)} configuration removes all proposed modules, training the task from scratch via TD-MPC2.

\textbf{Result.}
As shown in Tab.~\ref{tab:ablation_door} and Fig.~\ref{fig:ablation}, our \textbf{Full Model} achieves superior performance on the door-opening task (303.95 return, 12.64 w-steps), validating our integrated architecture. Removing the \textbf{intelligent router} (w/o Router) maintains high performance (296.57 return) but significantly increases training time (20.36 w-steps), confirming the router's critical role in efficient few-shot learning. Replacing VLM guidance with a rule-based router (w/o VLM) leads to complete task failure, demonstrating that VLM-provided semantic understanding is indispensable for goal-oriented compositional generalization. Without \textbf{imitation learning} (w/o IL), the model achieves moderate returns (193.61) but fails to converge ($\infty$ w-steps), proving that the IL stage provides a vital foundation for SEP-module compatibility. Finally, the \textbf{TD-MPC2 baseline} performs substantially worse (198.42 return, 32.38 w-steps), further underscoring the efficacy of our design.

\section{Conclusion and Future Works}
The MetaWorld-X is a hierarchical control framework that integrates world model representations, human motion priors, and semantic-driven expert composition for high-DoF humanoid control. By orchestrating prior-regularized experts via VLM-guided semantic routing, our approach mitigates skill interference while preserving biomechanical naturalness. 
Extensive experiments demonstrate significant improvements in training efficiency, task success rate, and motion quality over strong monolithic and model-based RL baselines. Our findings suggest that predictive modeling or imitation alone cannot resolve structural coupling; instead, semantically grounded modularity is essential for stable, natural, and generalizable behavior. 
Future work will explore tighter integration between semantic reasoning and dynamics modeling, as well as scaling the expert composition framework to larger skill libraries and real-world systems.

\vspace{0.1in}
{\small
\bibliographystyle{IEEEtran}
\bibliography{root}
}

\end{document}